\documentclass[conference]{IEEEtran}
\usepackage{graphicx}
\usepackage{float}
\usepackage{framed}
\usepackage{wrapfig}
\usepackage{enumerate}
\usepackage{algorithm}
\usepackage{algpseudocode}
\usepackage{caption}
\usepackage{subcaption}
\usepackage{adjustbox}
\usepackage{textcomp}
\usepackage{xcolor}
\usepackage{newtxmath,mathtools}
\usepackage{multirow}
\usepackage{array}
\usepackage{ifthen}
\usepackage{tabularx}
\usepackage{booktabs}
\usepackage{url}
\usepackage{bbm}
\usepackage[numbers]{natbib}
\usepackage{nameref}
\usepackage{makecell}

\usepackage{graphicx}
\usepackage{cancel}
\DeclareCaptionLabelFormat{figure}{Figure #2}
\usepackage{ifthen}
\usepackage{accessibility}
\newboolean{uselongversion}
\newcommand{\textversion}[2]{
  \ifthenelse{\boolean{uselongversion}}
    {#1}
    {#2}
    \unskip
}
\newcommand{\negspace}{\vspace{-0.50\baselineskip}}
\setboolean{uselongversion}{false}
\author{%
  Barak Gahtan, Robert J. Shahla, Reuven Cohen, Alex M. Bronstein \\
  Technion Israel Institute of Technology, Haifa, Israel \\
  \texttt{\{barakgahtan, shahlarobert, rcohen, bron\}@cs.technion.ac.il}
}

\begin{document}
\title{Estimating the Number of HTTP/3 Responses in QUIC Using Deep Learning}
\maketitle
\begin{abstract}
QUIC, a new and increasingly used transport protocol, enhances TCP by offering improved security, performance, and stream multiplexing.
These features, however, also impose challenges for network middle-boxes that need to monitor and analyze web traffic. This paper proposes a novel method to estimate the number of HTTP/3 responses in a given QUIC connection by an observer. This estimation reveals server behavior, client-server interactions, and data transmission efficiency, which is crucial for various applications such as designing a load balancing solution and detecting HTTP/3 flood attacks. The proposed scheme transforms QUIC connection traces into image sequences and uses machine learning (ML) models, guided by a tailored loss function, to predict response counts. Evaluations on more than seven million images—derived from 100,000 traces collected across 44,000 websites over four months—achieve up to 97\% accuracy in both known and unknown server settings and 92\% accuracy on previously unseen complete QUIC traces.
\end{abstract}

\section{Introduction}
Quick UDP Internet Connections (QUIC)~\cite{rfc9000} is increasingly replacing TCP as a primary transport protocol due to enhanced security, performance, and stream multiplexing. In HTTP/3~\cite{rfc9114}, QUIC can simultaneously deliver multiple HTTP objects over separate streams, reducing head-of-line blocking and improving responsiveness. Each stream’s data is carried in distinct QUIC frames, and each QUIC packet may contain multiple frames from different objects. Since streams operate independently, a delay on one stream does not impede others~\cite{rfc9114,rfc9308}. HTTP/3 maps HTTP semantics onto QUIC streams, assigning a client-to-server request to one stream and the corresponding server-to-client response to another~\cite{rfc9114}.

This paper considers an observer that sees packets flowing between a QUIC client and server, and aims to estimate how many HTTP objects the connection carries. Such estimation aids various tasks, including load balancing~\cite{shahla2024trafficgrinder}, where knowledge of how many requests are multiplexed within a single connection is crucial. Another use case is detecting HTTP/3 flood attacks~\cite{chatzoglou2023hands}, where multiple requests are sent in quick succession. Since the attack pattern closely resembles normal traffic, identifying it remains a challenge.

This paper introduces \emph{DecQUIC}, a scheme enabling a passive observer to estimate the number of request/response pairs in a QUIC connection. Building on prior work of~\cite{horowicz2022few,shapira2019flowpic}, we first capture a QUIC connection trace and divide it into multiple time windows. For each window, we generate two histograms: one for packets sent by the client and another for packets sent by the server. Combined with packet length, timing, and density information, these histograms form an RGB image, where the red channel encodes server packets, the green channel encodes client packets, and the blue channel is unused. Compared to the single-channel grayscale images in earlier works of~\cite{shapira2019flowpic}, our RGB representation provides directional and density cues essential for distinguishing concurrent HTTP/3 streams. Single-channel images offer only broad overviews, insufficient for parsing complex HTTP/3 traffic patterns within QUIC.

With the trace represented as a sequence of RGB images, we train an ML model to predict the number of HTTP/3 responses initiated in each image window. The scheme can also be evaluated for requests similarly. DecQUIC supports both online estimation (e.g., within 100\,ms of connection start) and offline assessment (e.g., over longer durations). Because the task involves discrete counts rather than class labels or continuous values, we define a discrete regression problem and develop a specialized loss function tailored to counting errors. 

To train and evaluate DecQUIC, we curated a labeled dataset of over 100,000 QUIC traces from 44,000 websites, collected across four months and multiple vantage points. From these traces, we generated over seven million images with varying window lengths. Using time windows of $T=0.1$ or $T=0.3$ seconds, DecQUIC achieved up to 97\% accuracy. \footnote{We make the code and the collected dataset available using \\ \url{https://github.com/robshahla/VisQUIC}.}

While counting HTTP/3 responses might appear simpler than comprehensive QUIC traffic classification, traditional non-ML methods can struggle with parallel HTTP/3 streams and overlapping packet patterns in a single QUIC connection. The ML-based approach in DecQUIC exploits fine-grained, directional, and density features in a scalable way, improving on hand-crafted heuristics by learning subtle temporal and length-based signatures. Moreover, as QUIC evolves and server behaviors diversify, ML models can adapt more flexibly than fixed rules, ensuring our method remains robust across a wide range of deployment scenarios.

The rest of this paper is organized as follows: Section~\ref{RELATEDSEC} reviews related work. Section~\ref{methodology} describes the proposed deep learning (DL) scheme and its challenges. Section~\ref{LossFunction} discusses the proposed loss function used for training the ML models. Section~\ref{eval-res} presents an evaluation of the trained ML models on out-of-training sample QUIC traces and out-of-distribution sample web servers. Section~\ref{tracesanalyzer} uses the trained ML models to estimate the number of HTTP/3 responses over complete QUIC traces. Finally, Section~\ref{conclusion} concludes the paper.

\section{Related Work}\label{RELATEDSEC}
The growth of encrypted network traffic has driven the increased use of ML techniques for flow-based analysis. Classical methods such as $k$-Nearest Neighbors, Random Forest, Naive Bayes, and Support Vector Machines~\cite{pacheco2020framework,sun2010novel,velan2015survey} have been widely applied, and more recently, deep learning (DL) approaches have emerged for classifying encrypted communications~\cite{lotfollahi2020deep,wang2018datanet}.

Several studies have employed CNN-based approaches for QUIC traffic classification. For instance, a two-stage method~\cite{tong2018novel} achieved high accuracy on QUIC-based services, while a multi-task approach~\cite{rezaei2020multitask} predicted various flow attributes and outperformed simpler models on the ISCX~\cite{ISCXDataset} and QUIC~\cite{QUICDataset} public datasets. However, these techniques were trained primarily on Google's services and focused on classifying service types rather than estimating connection characteristics as we do.

Ensemble-based ML methods have also been explored for QUIC classification~\cite{almuhammadi2023quic}, achieving high accuracy on a limited set of traffic types sourced solely from a single Google server. In contrast, our approach handles a broader set of web servers and a far larger dataset.
 
Other approaches have shown strong results under constrained conditions. For example, satellite-only QUIC analysis~\cite{secchi2022exploring} and a self-supervised method trained on Google-only traffic~\cite{towhid2022encrypted} both achieve high accuracy. These approaches, however, focus on specific environments or service-based classification, unlike our method, which estimates fundamental connection characteristics across diverse servers.

Other work has distinguished VPN from non-VPN encrypted traffic using complex ML pipelines~\cite{izadi2022network,izadi2022network11}, achieving high accuracy. However, this binary classification differs from our goal of estimating the number of HTTP/3 responses in diverse QUIC traffic.

CESNET-QUIC22~\cite{luxemburk2023cesnet} provides a large, diverse dataset of QUIC traffic, but it includes limited packet-level information and lacks HTTP/3 protocol details. These constraints make it unsuitable for tasks like ours, which require full-connection insights to estimate HTTP/3 responses.

Recent efforts using the CESNET-QUIC22 dataset have focused on classifying QUIC services~\cite{luxemburk2023encrypted,geiginger2021classification}, achieving fine-grained accuracy but not addressing out-of-distribution scenarios. In contrast, our work estimates connection characteristics rather than services and explicitly handles previously unseen servers.

In contrast to these existing classification efforts, our work addresses the subtler challenge of counting concurrent HTTP/3 responses in scenarios with highly imbalanced classes and partial concurrency. Most prior methods assume a limited server set (often Google-centric) and do not explicitly tackle out-of-distribution servers or the fine-grained concurrency within modern QUIC streams. By focusing on response counts, we enable tasks such as load balancing and attack detection, where concurrency plays a crucial role.

\section{DecQUIC Framework}\label{methodology}
We consider an observer monitoring encrypted QUIC traffic. Although payloads are hidden, the observer knows each packet's direction, length, and timestamp. Our goal is to estimate how many HTTP/3 responses occur within a connection by representing raw QUIC traces as labeled images suitable for ML.

\begin{figure}[t]
    \centering
    \includegraphics[width=0.45\textwidth]{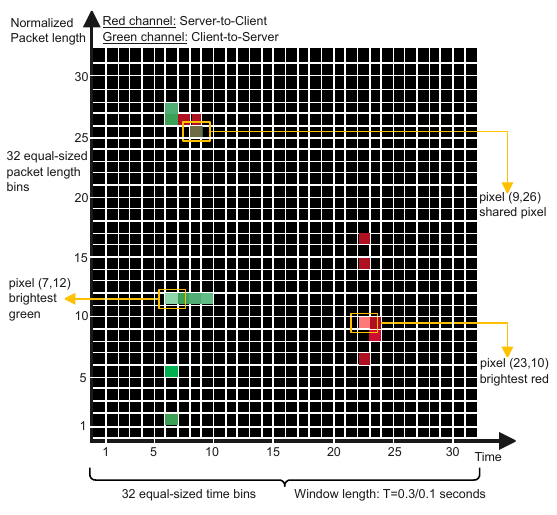}
    \caption{A DecQUIC image, representing QUIC connection activity. Pixel positions represent histogram bins (horizontal and vertical axes corresponding to time and packet length, respectively). The values of the red and green channels represent normalized, per-window, histogram counts of the response and request packets, respectively.}
    \label{fig:extended-miniflowpic}
\end{figure}
We segment each QUIC trace into fixed-length time windows ($T=0.1$ or $T=0.3$\ seconds) using a sliding window technique~\cite{frank2001time}. During training, we use a 90\% overlap between consecutive windows to increase data diversity. For evaluation, we use 0\% overlap, ensuring each response is counted once. By decrypting QUIC with available SSL keys, we determine how many HTTP/3 responses start within each window, thereby labeling each image.

Fig~\ref{fig:extended-miniflowpic} shows an example of the constructed image. Each window is divided into $M=32$ time bins and $N=32$ length bins, creating a $32\times32$ grid. We count packets per bin in both client-to-server (green channel) and server-to-client (red channel) directions, apply min-max normalization, and map the values to pixel intensities (0–255). The blue channel is unused. This approach extends FlowPic~\cite{shapira2019flowpic}, which only produced single-channel images, by incorporating directional and density information critical for multiplexed HTTP/3 streams. Additionally, balancing bin size is essential: coarser bins lose detail, while finer bins increase computation complexity with little accuracy gain~\cite{horowicz2022few}. Our tested values are 32, 64, 128 and 256, as was discussed in the work of ~\cite{horowicz2022few}. 
\begin{figure}[t]
    \centering
    \begin{subfigure}{0.15\textwidth}
        \centering
        \includegraphics[width=\textwidth]{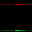}
        \caption{$2$ responses}
        \label{fig:2label}
    \end{subfigure}%
    \hspace{1mm}
    \begin{subfigure}{0.15\textwidth}
        \centering
        \includegraphics[width=\textwidth]{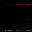}
        \caption{$6$ responses}
        \label{fig:6label}
    \end{subfigure}
    \hspace{1mm}
    \begin{subfigure}{0.15\textwidth}
        \centering
        \includegraphics[width=\textwidth]{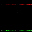}
        \caption{$12$ responses}
        \label{fig:12label}
    \end{subfigure}%
    \caption{Three DecQUIC images with their HTTP/3 response counts. Even visually similar images may differ in labels. \negspace \negspace \negspace}
    \label{fig:different-responses}
\end{figure}
\begin{figure}[t]
    \centering
    \includegraphics[width=0.5\textwidth]{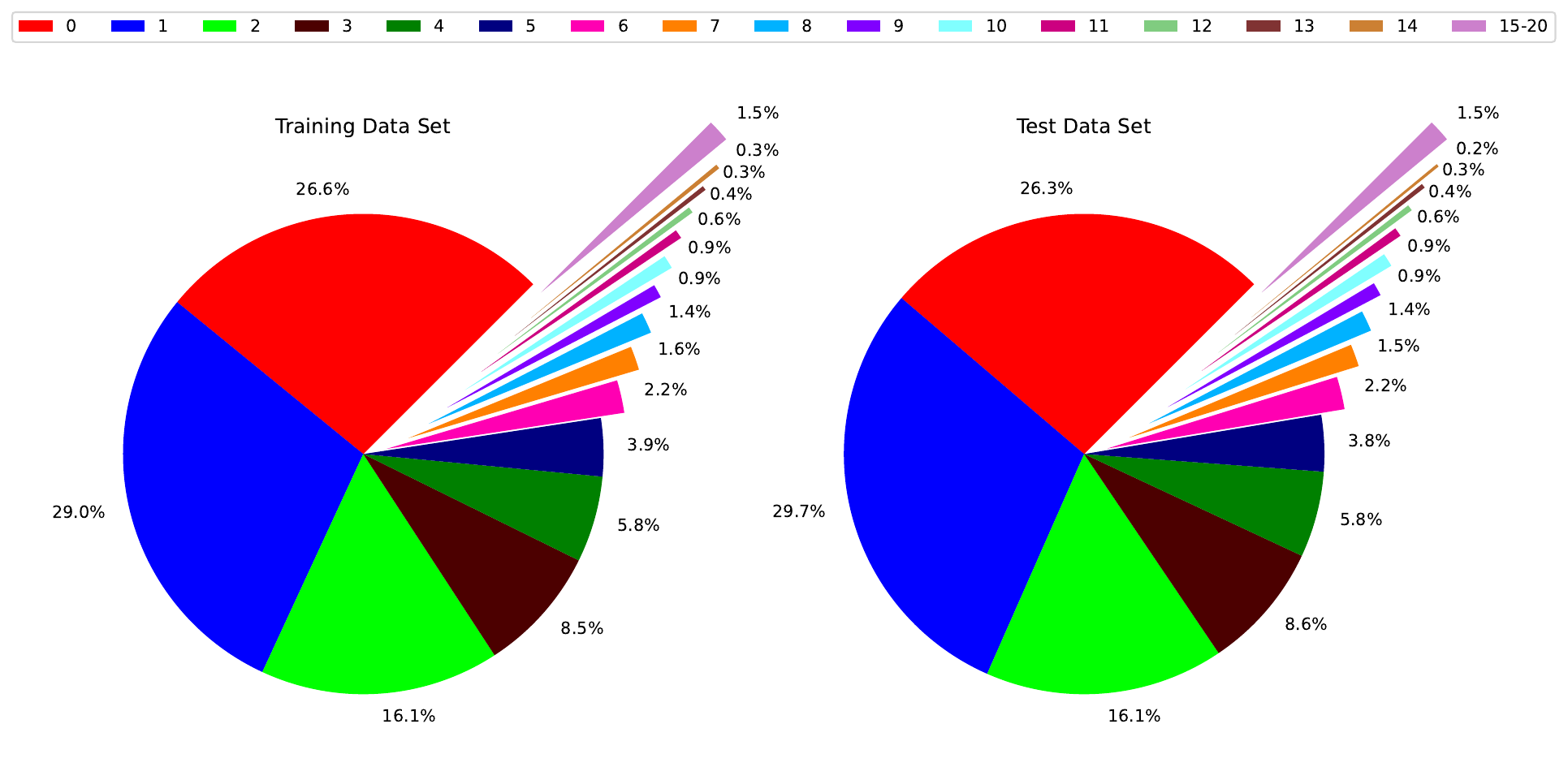}
    \caption{Response distribution for training and evaluation datasets with a $T=0.1$-second window.}
    \label{fig:response-distribution-dataset}
\end{figure}

Our dataset, derived from over 100,000 QUIC traces from 44,000 websites from tens of web servers captured from various vantage points, yields millions of images. However, several challenges arise. First, visually similar images can represent vastly different response counts (Fig~\ref{fig:different-responses}). Second, the distribution of classes (i.e., number of responses) is skewed, making some counts rare. We include only the distribution for the $T=0.1$-second dataset in Fig~\ref{fig:response-distribution-dataset} for brevity. The $T=0.3$-second dataset exhibits similar trends, with slight variations in the distribution due to the larger window size. Third, predicting an integer response count is a discrete regression task, not a standard classification or continuous regression scenario.

To address class imbalance, we apply minimal noise-based augmentation to underrepresented classes (those with a high response count above 10). Specifically, we add low-level Gaussian noise ($\sigma=2.55$, i.e., 1\% of the 8-bit pixel range) only to the minority-class images. This slight perturbation introduces subtle variability in pixel intensities, helping the model generalize better without disrupting temporal dependencies or altering the fundamental traffic patterns. We also design a custom loss function that penalizes large deviations between predicted and actual response counts. 
\begin{figure}[t!]
    \centering
    \includegraphics[width=1\linewidth]{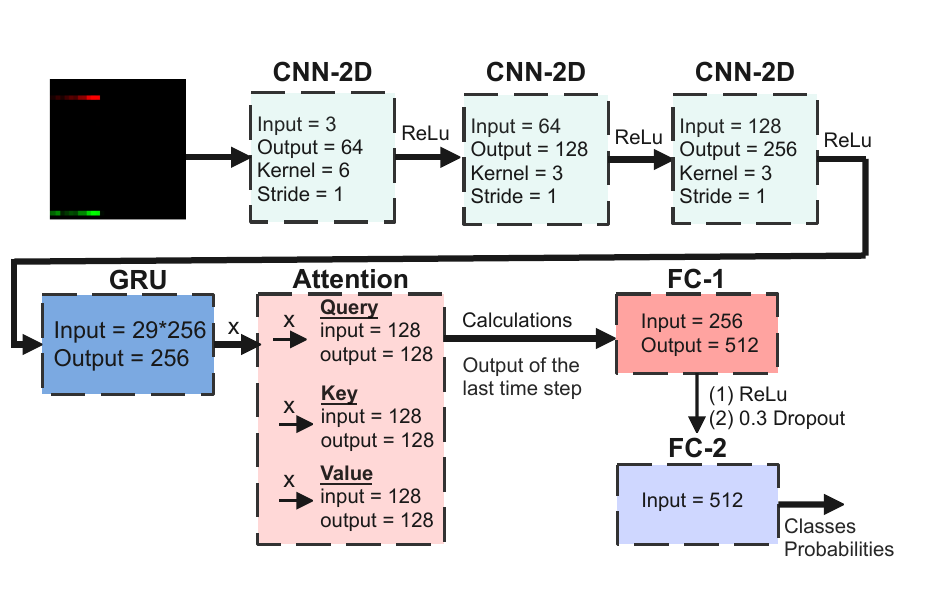}
    \caption{The proposed DecQUIC neural network architecture.}
    \label{fig:Neural-network-architecture}
\end{figure}

Our neural network architecture (Fig~\ref{fig:Neural-network-architecture}) is based on~\cite{ismail2020inceptiontime,khan2021attention}, and it combines CNN layers, for spatial feature extraction, a GRU, to capture temporal patterns, and a self-attention~\cite{vaswani2017attention} mechanism, to focus on critical time bins. Two fully connected layers finalize the prediction. Section~\ref{LossFunction} provides more details on the custom loss. Through the rest of the paper, the term ``class'' is used to denote the number of responses associated with each image (the label).

\section{A Discrete Regression Loss Function}\label{LossFunction}
During model development, we tested several standard loss functions, each targeting different aspects of our discrete, ordinal, and imbalanced prediction task. Cross-entropy struggled with class imbalance, while MSE and MAE—designed for continuous regression—overlooked the discrete and ordinal nature of our labels. The Huber loss proved robust to outliers but did not preserve class order.

These limitations led us to propose a new aggregated loss:
\begin{equation}
    L = \alpha\,\mathrm{FL} + (1-\alpha)\bigl(\beta\,\mathrm{ORL} + (1-\beta)\,\mathrm{DBL}\bigr).
\nonumber
\end{equation}
It combines three components: (1) a \emph{focused loss} (FL)~\cite{lin2017focal} to mitigate class imbalance by emphasizing hard-to-classify samples; (2) a \emph{distance-based loss} (DBL)~\cite{wang2020comprehensive} that penalizes predictions based on their distance from the true class; and (3) an \emph{ordinal regression loss} (ORL)~\cite{rennie2005loss, frank2001simple}, which maintains the natural order of the classes.

The FL term \cite{lin2017focal} builds on the weighted cross-entropy loss \cite{de2005tutorial} by adding a focusing parameter, \(\gamma \), which adjusts the influence of each sample on the training process based on the classification confidence. This parameter, \( \gamma \), modifies the loss function by scaling the loss associated with each sample by \((1 - p_t)^\gamma\), where \(p_t\) is the predicted probability of the true class \( \mathbf{y} \). This scaling reduces the loss from easy examples (where \( p_t \) is high), thereby increasing it for hard, misclassified examples, focusing training efforts on samples where improvement is most needed. Accordingly, the term is:
\begin{equation}
\mathrm{FL}(\mathbf{x}, \mathbf{y}) = \mathbb{E}_{(\mathbf{x}, \mathbf{y})} \left[ -w(y) \cdot \left(1 - \hat{\mathbf{y}}_{y}(\mathbf{x})\right)^\gamma \cdot \mathbf{y}^{\mathrm{T}}\log \hat{\mathbf{y}}(\mathbf{x}) \right],
\nonumber
\end{equation}
where \(\mathbf{x}\) denotes the input sample, \(\mathbf{y}\) is the one-hot encoded ground truth label, \(\hat{\mathbf{y}}(\mathbf{x})\) represents the model's output of class probabilities, \(\hat{\mathbf{y}}_{y}(\mathbf{x})\) denotes the predicted probability of the true class \(y\), and \(w(y)\) is a weight inversely proportional to the class frequency of \(y\) in the training dataset. By assigning a higher weight to less frequent classes, the model places more emphasis on accurately classifying these classes during training. It is an effective strategy for dealing with class imbalance~\cite{aurelio2019learning, tian2020recall, lin2017focal}. FL thus minimizes the relative loss for well-classified examples, while emphasizing difficult-to-classify ones. 

The DBL term~\cite{wang2020comprehensive}
\begin{equation}
\mathrm{DBL} = \mathbb{E}_{(\mathbf{x}, y)} \left[ 
\sum_{i} \hat{y}_i(\mathbf{x}) \cdot |i-y|
\right],
\nonumber
\end{equation}
with $y$ denoting the ground truth class,
is essentially a discrete regression loss that penalizes the model's output according to the predicted class's distance from the true label. The distance is computed as the absolute difference between the class indices and the target class. 

Finally, the ORL term \cite{rennie2005loss, frank2001simple} is given by
\begin{equation}
\mathrm{ORL} = \mathbb{E}_{(\mathbf{x}, \mathbf{y})} \left[ -\mathbf{y}^{\mathrm{T}}\log \sigma(\hat{\mathbf{y}}) - (1 - \mathbf{y})^{\mathrm{T}}\log \sigma(1-\hat{\mathbf{y}}) \right], \nonumber
\end{equation}
with $\sigma$ denoting the sigmoid function saturating the input between $0$ and $1$. ORL uses a binary cross-entropy loss function, which compares the activation of each output neuron to a target that shows if the true class is greater than or equal to each class index, thus helping the model determine the order of the classes. Both DBL and ORL consider the relations between classes; they do so in different ways: DBL penalizes predictions based on the numerical distance, while ORL makes explicit use of the classes' order. It focuses on preserving the correct order among predictions rather than the numerical distance between them.

Parameters \(\alpha\), \(\beta\), and \(\gamma\) control the relative influence of these components. Higher \(\alpha\) emphasizes FL, while lower \(\alpha\) favors ORL+DBL. A higher \(\beta\) increases the weight of ORL over DBL. Increasing \(\gamma\) within FL heightens focus on the hardest examples. In Section~\ref{eval-res}, we present an ablation study showing the benefits of each term.

\section{Evaluating the Machine Learning Models}\label{eval-res}
\begin{table}[!t]
\centering
\footnotesize
\caption{Summary statistics of QUIC traces and images per dataset of each web server.}
\footnotesize
\begin{tabular}{|c|c|c|c|c|}
\hline
\textbf{Web Server} & \textbf{Websites} & \textbf{Traces} & \textbf{$T=0.1$} & \textbf{$T=0.3$} \\ \hline
youtube.com & 399 & 2,109 & 139,889 & 54,659 \\ \hline
semrush.com & 1,785 & 9,489 & 474,716 & 221,477 \\ \hline
discord.com & 527 & 7,271 & 623,823 & 235,248 \\ \hline
instagram.com & 3 & 207 & 17,003 & 7,112 \\ \hline
mercedes-benz.com & 46 & 66 & 9,987 & 2,740 \\ \hline
bleacherreport.com & 1,798 & 8,497 & 781,915 & 331,530 \\ \hline
nicelocal.com & 1,744 & 1,666 & 148,254 & 48,900 \\ \hline
facebook.com & 13 & 672 & 25,919 & 10,988 \\ \hline
pcmag.com & 5,592 & 13,921 & 1,183,717 & 385,797 \\ \hline
logitech.com & 177 & 728 & 56,792 & 28,580 \\ \hline
google.com & 1,341 & 2,149 & 81,293 & 29,068 \\ \hline
cdnetworks.com & 902 & 2,275 & 207,604 & 85,707 \\ \hline
independent.co.uk & 3,340 & 3,453 & 176,768 & 68,480 \\ \hline
cloudflare.com & 26,738 & 44,700 & 1,347,766 & 341,488 \\ \hline
jetbrains.com & 35 & 1,096 & 34,934 & 18,470 \\ \hline
pinterest.com & 43 & 238 & 6,465 & 2,360 \\ \hline
wiggle.com & 4 & 0 & 0 & 0 \\ \hline
cnn.com & 27 & 2,127 & 91,321 & 59,671 \\ \hline
\end{tabular}
\label{tab:combined-table}
\end{table}
We now present a quantitative evaluation of our framework under two conditions: known and unknown web servers. In the former case, a set of models were trained and evaluated exclusively on the QUIC traces pertaining to the web servers assumed at inference time. In the latter case, a leave-two-servers-out evaluation was performed. We use a leave-two-servers-out approach—chosen because we have 18 web servers and aim to reserve about 10\% of the web servers for out-of-distribution evaluation. Details on training and test set construction are provided later in this section.

Models were trained with both $T=0.1$ and $T=0.3$ second windows on traces from 44,000 websites and 100,000 traces collected over four months from 18 different web servers that support QUIC. Table~\ref{tab:combined-table} summarizes server-level statistics. Classes above 20 are exceedingly rare (e.g., only 0.003\% of images are labeled as class 21) and thus excluded from training and testing, leaving the majority (90\% for $T=0.1$\ seconds and 95\% for $T=0.1$\ seconds windows) concentrated on classes 0–20.
\subsection{Results for Known Web Servers}
We first evaluate our framework when the web servers are known. Each server’s traces were split $80:20$ into training and testing sets, and we trained five models on different random splits. During training we used a batch size of 64, the Adam optimizer, and a ReduceLROnPlateau scheduler that reduced the learning rate by 30\% upon reaching a validation-loss plateau. Early stopping with a patience of six epochs prevented overfitting. We performed a grid search to determine the optimal loss parameters $\alpha,\beta,\gamma$ by selecting the combination that yielded the lowest validation loss. We tested $\alpha \in \{0,0.3,0.5,0.7,1\}$, $\beta \in \{0,0.4,0.6,1\}$, and $\gamma \in \{1,2,3\}$. For $T=0.1$\ second windows, $\alpha=0.7$, $\beta=0.4$, and $\gamma=2$ were optimal; for $T=0.1$\ second windows, $\gamma=3$ worked best with the same $\alpha$ and $\beta$.
\begin{figure*}[t!]
    \centering
    \begin{subfigure}[b]{0.45\textwidth}
        \centering
        \includegraphics[width=\textwidth]{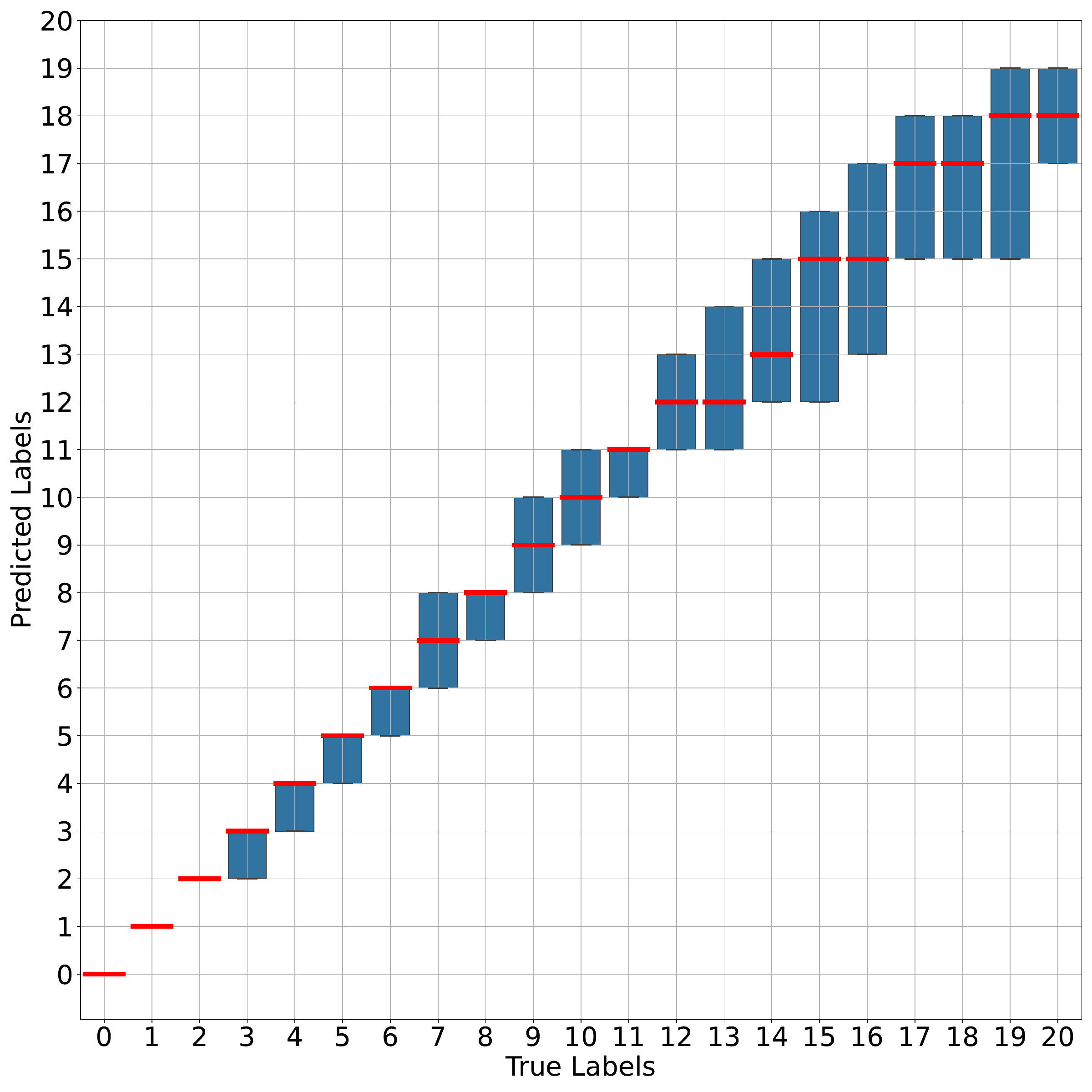}
        \caption{Window length $T=0.1$ second}
        \label{fig4:boxplot01mix}
    \end{subfigure}%
    \begin{subfigure}[b]{0.45\textwidth}
        \centering
        \includegraphics[width=\textwidth]{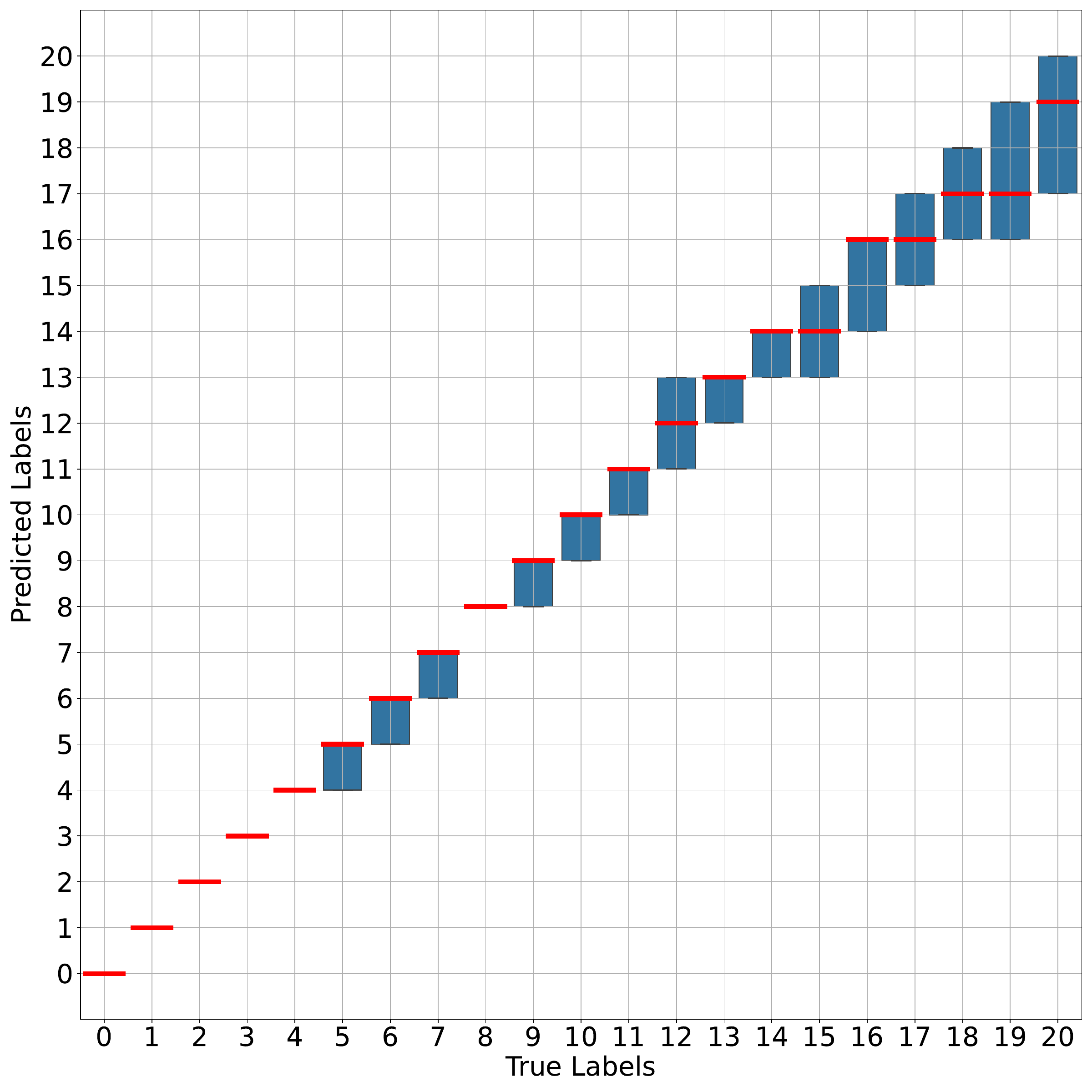}
        \caption{Window length $T=0.3$ second}
        \label{fig4:boxplot03mix}
    \end{subfigure}
    \caption{Known-server prediction errors (iteration 1). Red lines: median; blue boxes: 25–75\% intervals.}
    \label{fig:box_plot_known}
\end{figure*}

Fig~\ref{fig:box_plot_known} shows box plots of predicted classes vs. true classes (0–20) for out-of-training-sample traces from one of the five iterations. For $T=0.1$\ second, lower-value classes ($0,1,2$) form thin lines, indicating minimal prediction variance and very high accuracy. As true class values increase, predictions spread more widely. For $T=0.3$ second, accurate predictions extend up to class $4$, and even the upper classes ($16$–$20$) show tighter, less biased distributions. This improvement aligns with dataset distributions: in the $T=0.1$\ seconds dataset, classes $0,1,2$ constitute about $75\%$ of the data, whereas in the $T=0.3$\ second dataset these classes form only $47\%$, yielding a more balanced scenario. Thus, $T=0.3$ second windows provide more robust standalone accuracy, advantageous for online use cases.
\begin{table}[t!]
\centering
\footnotesize
\caption{CAP results for known web servers, from five random training/test splits at $T=0.1$ and $T=0.3$.}
\begin{tabular}{|c|c|c|c|c|}
\hline
\multicolumn{1}{|c|}{\textbf{Iteration}} & \multicolumn{2}{c|}{\textbf{$T=0.1$}} & \multicolumn{2}{c|}{\textbf{$T=0.3$}} \\ \hline
\multicolumn{1}{|c|}{} & \textbf{$\pm$1} & \textbf{$\pm$2} & \textbf{$\pm$1} & \textbf{$\pm$2} \\ \hline
1 & 0.93 & 0.97 & 0.91 & 0.96 \\ \hline
2 & 0.92 & 0.96 & 0.90 & 0.97 \\ \hline
3 & 0.93 & 0.98 & 0.91 & 0.95 \\ \hline
4 & 0.94 & 0.97 & 0.92 & 0.93 \\ \hline
5 & 0.91 & 0.96 & 0.92 & 0.94 \\ \hline
\end{tabular}
\label{tab:CAP results combined known webservers.}
\end{table}

We also introduce the Cumulative Accuracy Profile (CAP), which tolerates a $\pm k$-class error:\\
$\mathrm{CAP}_{\pm k}(\mathbf{y},\hat{\mathbf{y}}) = \frac{1}{n}\sum_{i=1}^n \mathbbm{1}(|y_i-\hat{y}_i|\leq k),$ where \( \mathbf{y} \) represents the vector of true class labels, \( \hat{\mathbf{y}} \) denotes the model’s predictions, \( k \) specifies the tolerance level (e.g., \(\pm 1\) or \(\pm 2\) classes), \( n \) is the total number of samples, and \( \mathbbm{1}(\cdot) \) is the indicator function that evaluates to $1$ if the condition is met and $0$ otherwise.
Unlike exact-match metrics, CAP credits predictions close to the true label. Table~\ref{tab:CAP results combined known webservers.} presents CAP scores for five training/test splits, using both $T=0.1$\ seconds and $T=0.3$\ seconds datasets, without stratified sampling. The results show that a large majority of predictions fall within one or two classes of the correct label, reflecting strong performance.
\subsection{Results for Unknown Web Servers}
We now evaluate the model’s ability to generalize to web servers unseen during training. Here, we partition the datasets into ten iterations, each time holding out two web servers for testing and using the remaining servers for training. This leave-two-servers-out approach tests out-of-distribution performance, as client--server dynamics can differ widely across servers. For example, if ``semrush.com'' is held out, the model may never encounter high-value classes during training, making generalization on those classes difficult. Similarly, if testing servers lack higher-value classes (e.g., ``instagram.com'' or ``pcmag.com''), the model cannot learn to predict those classes at all. All models here use the same training procedure described previously.
\begin{figure}[t!]
    \centering
    \includegraphics[width=0.9\linewidth]{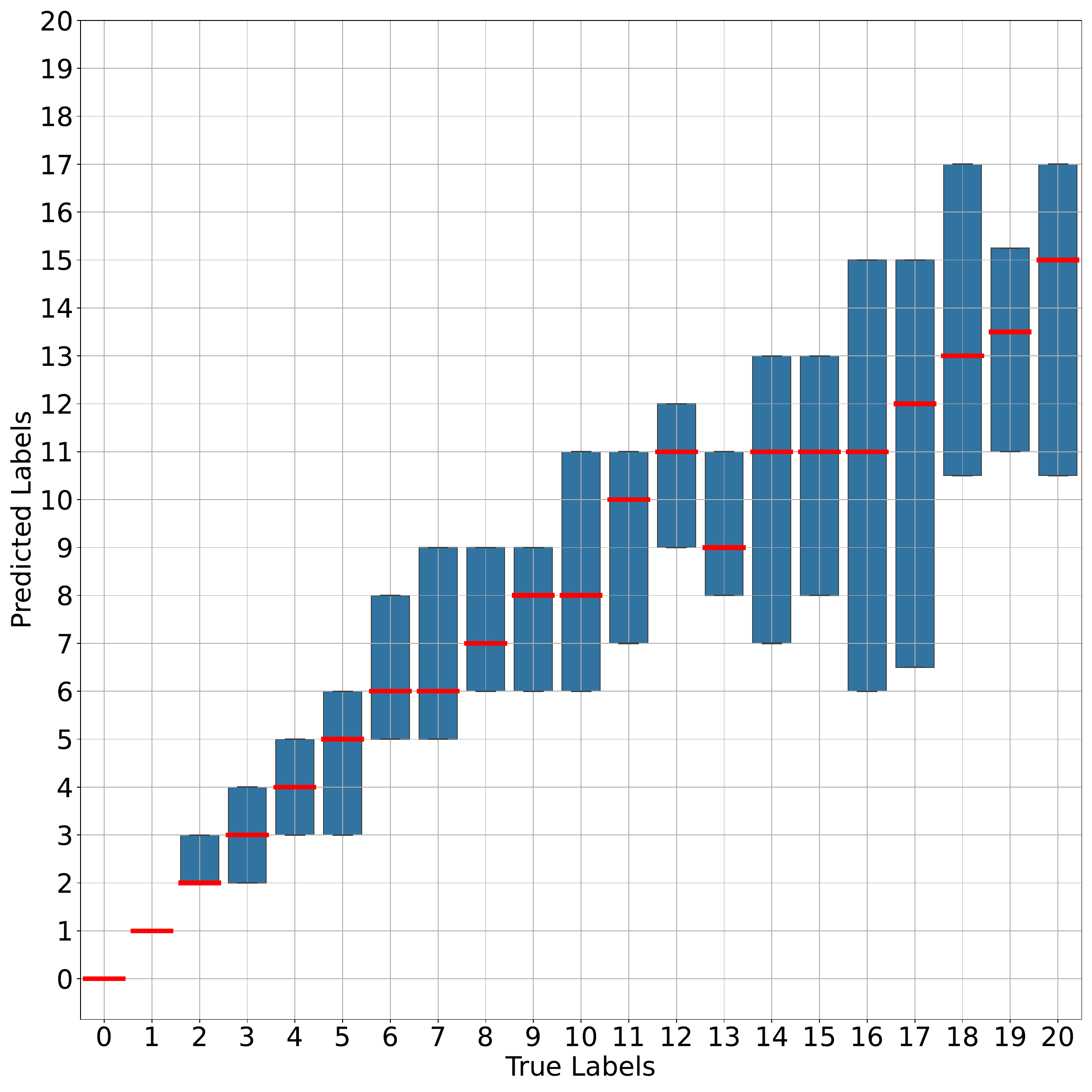}
    \caption{Unknown-server prediction errors ($T=0.1$) with independent.co.uk'' and google.com'' as test servers. Red: median; blue boxes: 25–75\% intervals. }
    \label{fig:0.3 sliding window length, unknown servers, all 21 classes boxplot and confusion}
\end{figure}
Fig~\ref{fig:0.3 sliding window length, unknown servers, all 21 classes boxplot and confusion} shows an example for $T=0.1$\ seconds where ``independent.co.uk'' and ``google.com'' are the testing servers. Predictions are accurate for lower classes $(0\!-\!1)$, remain within an acceptable $\pm2$ class range for mid-level classes $(2\!-\!12)$, and grow more scattered for higher-value classes. This iteration’s CAP is 83\% at $\pm1$ tolerance and 90\% at $\pm2$, illustrating partial generalization but reduced performance at upper classes.

\begin{table}[t!]
\centering
\footnotesize
\caption{CAP results (unknown servers) for 10 random iterations, each testing on 2 servers and training on the rest, using $T=0.3$-second windows.}
\begin{tabular}{|l|c|c|}
\hline
\multicolumn{1}{|c|}{\textbf{Testing Servers}} & \multicolumn{1}{c|}{\textbf{$\pm$1}} & \multicolumn{1}{c|}{\textbf{$\pm$2}} \\ \hline
bleacherreport.com, cloudflare.com & 0.62 & 0.76 \\ \hline
facebook.com, cdnetworks.com       & 0.59 & 0.72 \\ \hline
logitech.com, mercedes-benz.com    & 0.66 & 0.77 \\ \hline
bleacherreport.com, semrush.com    & 0.63 & 0.75 \\ \hline
independent.co.uk, google.com      & 0.83 & 0.90 \\ \hline
cnn.com, facebook.com              & 0.82 & 0.88 \\ \hline
discord.com, youtube.com           & 0.66 & 0.84 \\ \hline
discord.com, google.com            & 0.61 & 0.80 \\ \hline
discord.com, independent.co.uk     & 0.83 & 0.88 \\ \hline
bleacherreport.com, google.com     & 0.78 & 0.85 \\ \hline
\end{tabular}
\label{tab:CAP results 0.3 unknown webservers.}
\end{table}

\begin{table}[t!]
\centering
\footnotesize
\caption{CAP results (unknown servers) for 10 random iterations, each testing on 2 servers and training on the rest, using $T=0.1$-second windows.}
\begin{tabular}{|l|c|c|}
\hline
\multicolumn{1}{|c|}{\textbf{Testing Servers}} & \multicolumn{1}{c|}{\textbf{$\pm$1}} & \multicolumn{1}{c|}{\textbf{$\pm$2}} \\ \hline
jetbrains.com, semrush.com & 0.69 & 0.78 \\ \hline 
pcmag.com, discord.com       & 0.86 & 0.94 \\ \hline 
instagram.com, cloudflare.com    & 0.79 & 0.90 \\ \hline
instagram.com, bleacherreport.com    & 0.78 & 0.87 \\ \hline
youtube.com, jetbrains.com     & 0.86 & 0.92 \\ \hline 
pcmag.com, cloudflare.com              & 0.80 & 0.89 \\ \hline
facebook.com, nicelocal.com          & 0.75 & 0.85 \\ \hline
cdnetworks.com, independent.co.uk            & 0.71 & 0.81 \\ \hline
cnn.com, facebook.com     & 0.86 & 0.90 \\ \hline 
youtube.com, nicelocal.com     & 0.81 & 0.87 \\ \hline
\end{tabular}
\label{tab:CAP results 0.1 unknown webservers.}
\end{table}

Tables~\ref{tab:CAP results 0.3 unknown webservers.} and ~\ref{tab:CAP results 0.1 unknown webservers.} summarize CAP results for all ten iterations with $T=0.1$\ seconds and $T=0.3$\ seconds windows, respectively. While low-value classes remain well-predicted, mid and high-value classes pose greater challenges in the unknown-server setting.

\section{Evaluating on Complete Traces}\label{tracesanalyzer}
\begin{figure*}[t]
    \centering
    \begin{subfigure}[t]{0.9\columnwidth}
        \centering
        \includegraphics[width=\textwidth]{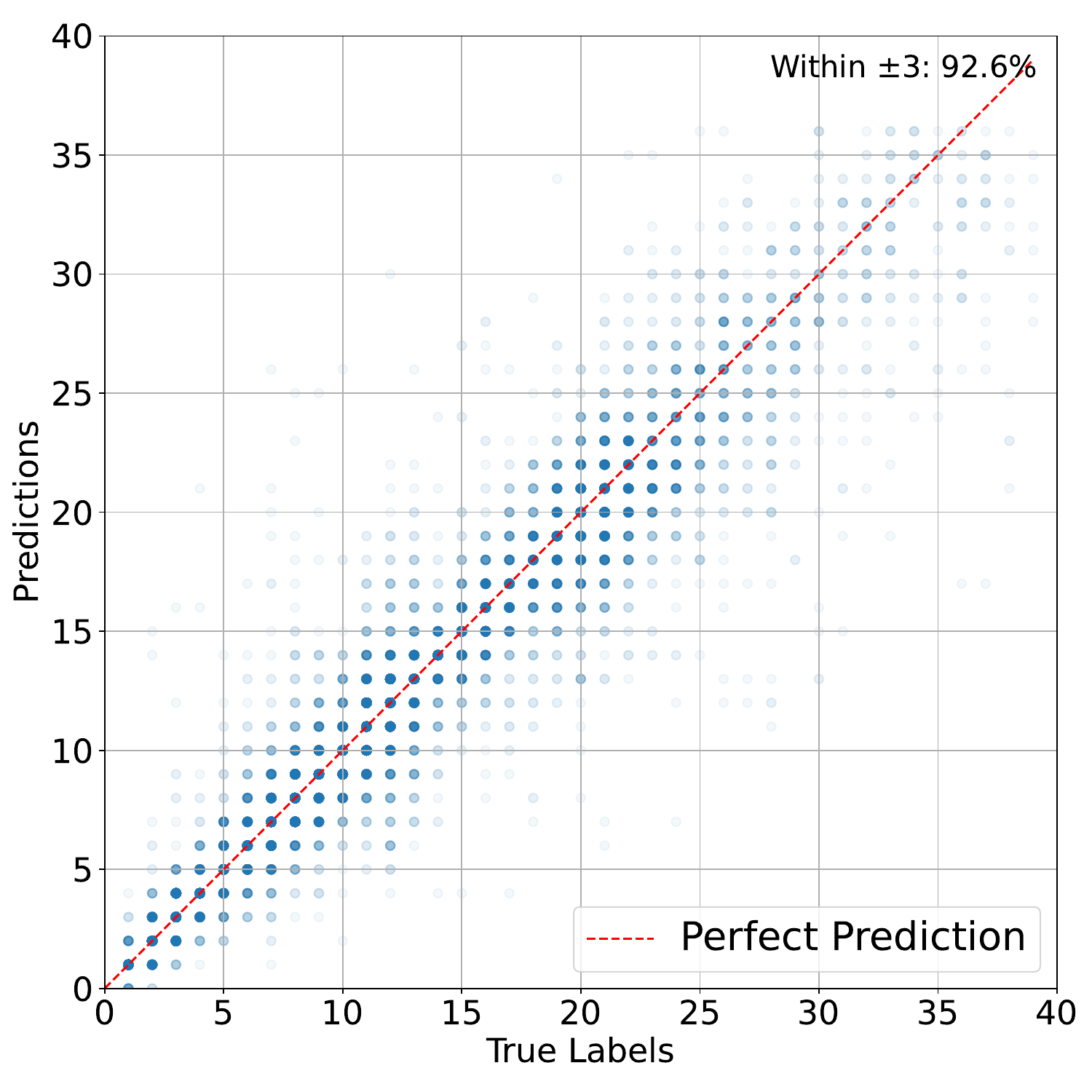}
        \caption{Window length $T=0.1$ second}
        \label{fig4:scatter_trace01}
    \end{subfigure}
    \begin{subfigure}[t]{0.9\columnwidth}
        \centering
        \includegraphics[width=\textwidth]{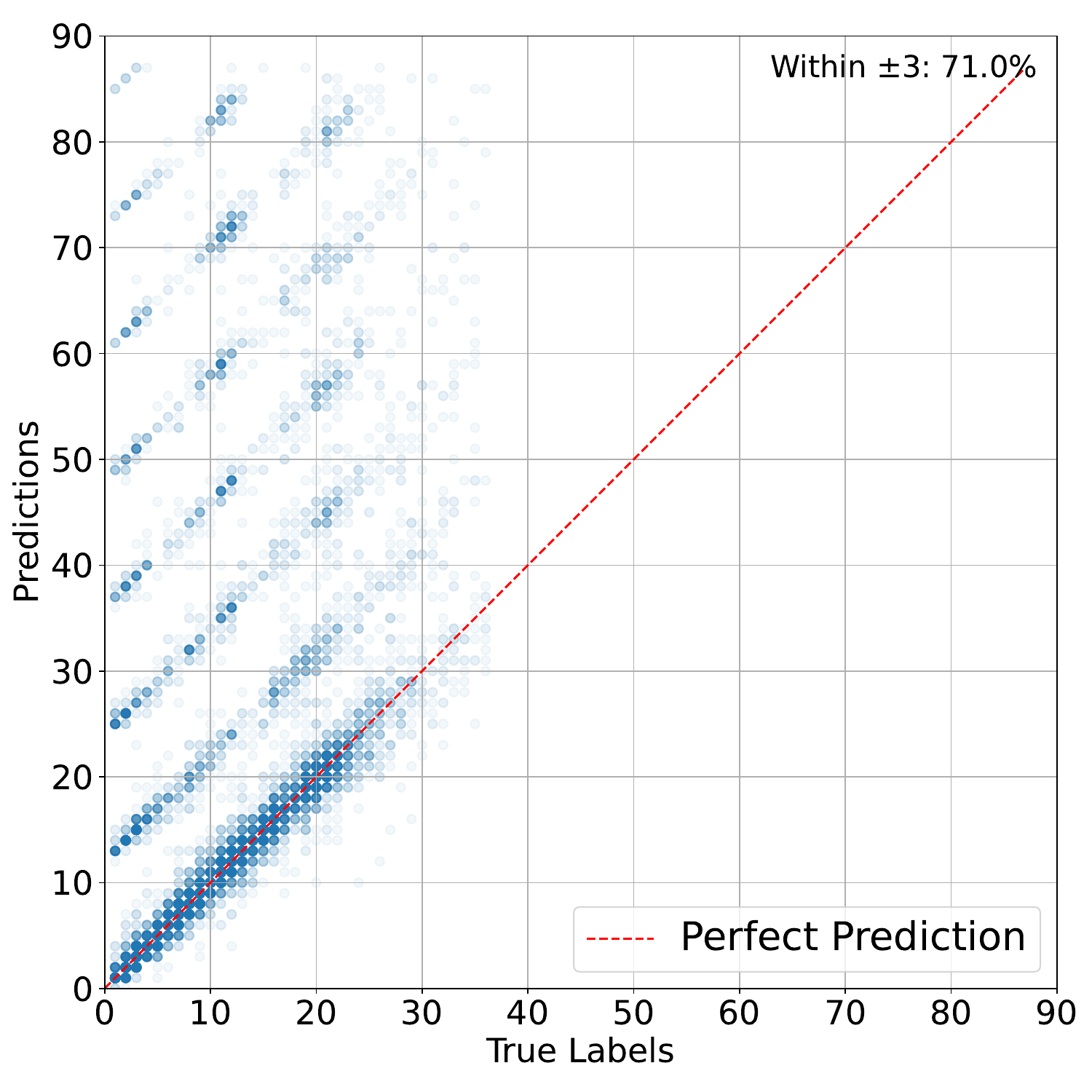}
        \caption{Window length $T=0.3$ second}
        \label{fig4:scatter_trace03}
    \end{subfigure}
    \caption{Scatter plots of summed predictions vs. true labels (transparency = 0.05) to highlight point density in overlapping regions. }
    \label{fig:traces-eval-images-03-model-03and01}
\end{figure*}

This section evaluates the model’s ability to estimate the total number of HTTP/3 responses in a trace. To do so, \textbf{we generated new, non-overlapping windows} from the same test traces used previously (Section~\ref{eval-res}, iteration 1, known-server scenario). We then summed the models’ per-window predictions and compared them to the true total responses per trace.

Fig~\ref{fig:traces-eval-images-03-model-03and01} presents scatter plots of predicted vs. true total responses for $T=0.1$\ seconds and $T=0.3$\,s windows. Each point represents one trace, and transparency ($\theta=0.05$) reveals areas of high point density. For example, if a trace is composed of five non-overlapping images with labels $1,0,2,4,1$ (total $8$) and model predictions $1,0,3,4,1$ (total $9$), it appears as $(8,9)$. Multiple traces with the same totals stack, increasing point opacity. 

For $T=0.1$\ seconds windows, the test set has 12,520 traces (avg. 21.2 images/trace); for $T=0.3$\,s windows, 12,142 traces (avg. 7.5 images/trace) are used. We use a $\pm3$ tolerance level because for both window lengths, the points represent the aggregated prediction sum and, thus, the aggregated errors as well. The $T=0.1$\ seconds model achieves 92.6\% accuracy versus 71\% for $T=0.3$\,s. Additionally, the $T=0.1$\ seconds predictions cluster more tightly along the diagonal, suggesting finer temporal granularity aids cumulative accuracy.

Figures~\ref{fig:traces-eval-images-03-model-03and01}(\subref{fig4:scatter_trace01}) 
and~\ref{fig:traces-eval-images-03-model-03and01}(\subref{fig4:scatter_trace03}) illustrate a notable difference in predictive behavior between the models trained and evaluated with $T=0.1$- and $T=0.3$-second window sizes. In particular, the $T=0.3$ model displays a series of strong diagonal patterns, whereas these are far less evident for the $T=0.1$ model. Several factors drive this phenomenon:
\begin{enumerate}
    \item \textbf{Variance in true labels:} 
    When using $T=0.1$-second windows, the majority of images fall into lower-valued classes (e.g., $0\!-\!3$ responses). Consequently, the model trained on such a distribution excels in this narrow lower-range. By contrast, a $T=0.3$-second window lumps together more concurrent requests, producing a broader range of labeled classes (including mid-to-high values). As a result, any systematic errors the model makes on higher counts tend to repeat across many samples, appearing as clustered diagonals in the scatter plot.
    \item \textbf{Cumulative effect of mispredictions:} 
    When summing per-window predictions to obtain the total number of responses per trace, even minor window-level misestimates can compound. If a single window is misestimated by a few counts (especially in mid-to-high classes), subsequent partial errors accumulate, shifting the aggregated prediction upward or downward. This systematic drift manifests visually as diagonal streaks above or below the perfect-prediction line.
    \item \textbf{Class imbalance at high values:}
    Although $T=0.3$-second windows increase the likelihood of higher-class labels, those classes are still relatively rare compared to classes near the mean. The model’s exposure to fewer high-class examples during training can lead to amplified inaccuracies on that subset, reinforcing the diagonal banding once these errors propagate across entire traces.
    \item \textbf{Coarser temporal granularity:}
    Finally, with larger windows, nuanced changes in packet arrival rates are “averaged out” in each image, potentially masking fine-grained cues that help distinguish, say, 10 from 12 concurrent responses. In contrast, $T=0.1$-second windows offer more granular snapshots, which, while increasing computational cost, can bolster per-trace accuracy by limiting each misestimation’s scope.
\end{enumerate}

Both figures highlight that the choice of window length is inherently use-case-dependent. Shorter windows typically enhance aggregated trace-level accuracy—particularly for offline analyses that require precise counts—whereas longer windows can yield better single-image accuracy (as seen in Section~\ref{eval-res}). However, the finer temporal resolution of $T=0.1$ seconds also increases computational overhead during both training and inference. Ultimately, balancing these trade-offs is essential for practitioners deploying DecQUIC in diverse QUIC environments.

\section{Conclusion}\label{conclusion}
We addressed the challenge of estimating the number of HTTP/3 responses in QUIC connections, a task essential for network management, load balancing, and service optimization. Despite QUIC’s encryption and variable conditions, our deep learning-based approach achieves high accuracy. Using a dataset of over seven million images from 100,000 QUIC traces, we evaluated models in both known and unknown server scenarios, reaching up to 97\% CAP accuracy. Moreover, we accurately estimated the total responses in over 12,000 traces with 92.6\% accuracy, demonstrating our method’s robustness and applicability. \negspace    

\begin{scriptsize}
\bibliographystyle{IEEEtran}{00}
\bibliography{refs}

\end{scriptsize}
\end{document}